\documentclass[lettersize,journal]{IEEEtran}
\usepackage{amsmath}
\usepackage{amsfonts,bm}
\usepackage{amsthm}
\usepackage{algorithm,algorithmicx,algpseudocode}
\usepackage{array}
\usepackage{textcomp}
\usepackage{stfloats}
\usepackage{url}
\usepackage{verbatim}
\usepackage{graphicx}
\usepackage{cite}
\usepackage{multirow}
\usepackage{booktabs}
\usepackage{hyperref}
\usepackage{stmaryrd}
\usepackage{amssymb}
\usepackage{pifont}
\usepackage{subcaption}
\usepackage{boldline}
\usepackage{cite}
\usepackage{enumitem}
\usepackage{calc}
\usepackage{multibib}

\usepackage{xcolor}
\newcommand{\tb}[1]{\textcolor{black}{#1}}

\newcommand{\cmark}{\ding{51}}%
\newcommand{\xmark}{\ding{55}}%
\newcommand{\compact}{{\sc Compact}}
\newcommand{\CL}{{\sc Compact Learning}}

\newcommand{\bb}[1]{\textbf{#1}}

\newcommand{\RR}{\mathbb{R}}

\newcommand{\cE}{\mathcal{E}}
\newcommand{\cG}{\mathcal{G}}
\newcommand{\cN}{\mathcal{N}}
\newcommand{\cL}{\mathcal{L}}

\newcommand{\Minimize}{\operatornamewithlimits{\text{\textbf{Minimize}}}}

\newcommand{\dataset}{\mathcal{D}}

\newcommand{\pd}{p^d}
\newcommand{\qd}{q^d}

\newcommand{\pg}{p^g}
\newcommand{\qg}{q^g}
\newcommand{\pgi}{p_i^g}
\newcommand{\qgi}{q_i^g}
\newcommand{\pgimin}{\underline{p}_i^g}
\newcommand{\pgimax}{\overline{p}_i^g}
\newcommand{\qgimin}{\underline{q}_i^g}
\newcommand{\qgimax}{\overline{q}_i^g}

\newcommand{\vm}{v}
\newcommand{\va}{\theta}
\newcommand{\vai}{\theta_i}
\newcommand{\vaj}{\theta_j}

\newcommand{\norm}[1]{\left\lVert#1\right\rVert} 

\newfloat{model}{thp}{lop}
\floatname{model}{Model}

\begin{document}

\title{Compact Optimization Learning for AC Optimal Power Flow}

\author{Seonho~Park, Wenbo~Chen, Terrence~W.K.~Mak, and Pascal~Van~Hentenryck
\thanks{
    The authors are affiliated with the School of Industrial and Systems Engineering, Georgia Institute of Technology, Atlanta, GA 30332, USA, 
    E-mail: \{seonho.park,wenbo.chen,wmak,pvh\}@gatech.edu
}
}

\maketitle

\begin{abstract}
This paper reconsiders end-to-end learning approaches to the Optimal
Power Flow (OPF). Existing methods, which learn the input/output
mapping of the OPF, suffer from scalability issues due to the high
dimensionality of the output space. This paper first shows that the
space of optimal solutions can be significantly compressed using
principal component analysis (PCA). It then proposes \CL{}, a new method
that learns in a subspace of the principal components and
translates the vectors into the original output space. This
compression reduces the number of trainable parameters substantially,
improving scalability and effectiveness. 
\CL{} is evaluated on a variety of test cases from the PGLib
and a realistic French transmission 
system having renewable energy changes with up to 30,000 buses. The
paper also shows that the output of \CL{} can be used to warm-start an
exact AC solver to restore feasibility, while bringing significant
speed-ups.
\end{abstract}


\section{Introduction}
\label{sec:intro}

Optimal Power Flow (OPF) is at the core of grid operations: in many
markets, it should be solved every five minutes (ideally) to clear
real-time markets at minimal cost, while ensuring that the load and
generation are balanced and that the physical and engineering
constraints are satisfied. Unfortunately, the AC-OPF problem is
nonlinear and nonconvex: actual operations typically use linear
relaxations (e.g., the so-called DC-model) to meet the real-time
requirements of existing markets.

As the share of renewable energies significantly increases in the
generation mix, it becomes increasingly important to solve AC-OPF
problems, not their linearized versions. This is exemplified by
the ARPA-E GO competitions \cite{safdarian2022grid} designed to 
stimulate progress on AC-OPF.
In recent years, machine-learning (ML) based approaches to the OPF have
received increasing attention. This is especially true for end-to-end
approaches that aim at approximating the mapping between various input 
configurations and corresponding optimal solutions
\cite{zamzam2020learning,fioretto2020predicting,donon2020deep,diehl2019warm,Owerko2020,nellikkath2022physics,donti2021dc3,park2022self}.
The ML approach is motivated by the recognition that OPF
problems are solved repeatedly every day, producing a wealth of
historical data. In addition, the historical data can be augmented with
additional AC-OPF instances, moving the computational burden offline
instead of solving during real-time operations.

One of the challenges of AC-OPF is the high dimensionality of its
solution, which implies that the ML models, typically
deep neural networks (DNNs), have an excessive number of trainable parameters
for realistic power grids. As a result, many ML
approaches are only evaluated on test cases of limited sizes. For
instance, the pioneering works in \cite{zamzam2020learning} and
\cite{fioretto2020predicting} tested their approaches on systems with
118 and 300 buses at most, respectively. To the best of the authors'
knowledge, the largest AC-OPF test case for evaluating end-to-end
learning of AC-OPF is the French system with around 6,700 buses
\cite{chatzos2021spatial}. Observe also that, since end-to-end
learning is a regression task, learning highly dimensional OPF output
may lead to inaccurate predictions and significant constraint
violations.

\tb{ To remedy this limitation and learn OPF at scale, this paper
  proposes a different approach.  Instead of directly mapping the
  inputs to the AC-OPF solutions, this paper proposes to learn a
  mapping to a low-rank representation defined by a Principal
  Component Analysis (PCA) before translating the vectors into the
  original output space. The contributions of the paper can be
  summarized as follows: }
\begin{itemize} 
\item \tb{A data analysis on the optimal solutions given
  various input configurations shows that the optimal solutions to the
  AC-OPF problems can be substantially compressed through PCA with
  negligible informational loss.}

\item \tb{Motivated by this empirical observation, the paper proposes
  \CL{}, a new ML approach that learns into a subspace of
  principal components before translating the compressed output into
  the original output space. In fact, \CL{} jointly learns both the
  selected principal components and the compact mapping function.}
  
\item \tb{\CL{} learns the AC-OPF mapping for very large power systems
  with up to 30,000 buses. To the best of the authors' knowledge, this
  is the largest AC-OPF problem to which an end-to-end learning scheme
  has been applied. The results show that \CL{} is comparable in
  accuracy to the best previous approach, but scales significantly
  better. }

\item \tb{The paper also demonstrates that the \CL{} predictions can 
  warm-start power flow and AC-OPF solvers. When seeding a power flow
  solver, \CL{} exhibits compelling performance compared with
  plain approaches. The warm-start results, which use both
  primal and dual predictions, show that \CL{} can produce significant
  speed-ups for AC-OPF solvers, which can be accelerated by a factor of
  $3.6$ to $15.3$\texttimes{} on the bulk industry size power systems. }
\end{itemize}

\noindent
The rest of this paper is organized as
follows: Section~\ref{sec:related} presents prior related works.
Section~\ref{sec:acopf_ml_based} revisits the AC-OPF formulation and
describes the supervised learning task. Section~\ref{sec:prelim}
analyzes the structure of optimal solutions. Section~\ref{sec:method}
presents \CL{} in detail. Section~\ref{sec:exp} demonstrates its
performance for various AC-OPF test cases. Finally,
Section~\ref{sec:conclusions} covers the concluding remarks.

\section{Related Work}
\label{sec:related}

The \emph{end-to-end learning} approach aims at training a model that
directly estimates the optimal solution to an optimization problem
given various input configurations. This approach has attracted significant
attention in power system applications recently because it holds the
promise of decreasing the computation time needed to solve recurring
optimization problems with reasonably small variations of the input
parameters. For example, a \emph{classification-then-regression} framework 
\cite{chen2022learning} was proposed to directly estimate
the optimal solutions to the security-constrained economic dispatch
problem. Because of the existence of the bound constraints, they
recognized that the majority of the generators are at their
maximum/minimum limits in optimal solutions. This study observed a
similar pattern in AC-OPF solutions, but a more efficient
way to design the input/output mapping is proposed.  
Similarly, ML-based mappings have been utilized to approximate the 
optimal commitments in various unit commitment problems
\cite{xavier2021learning,ramesh2022feasibility,park2022confidence}.
Especially for AC-OPF problems, various supervised learning (e.g.,
\cite{zamzam2020learning,fioretto2020predicting}) and self-supervised
learning approaches (e.g., \cite{park2022self,donti2021dc3}) have been
researched. They have used dedicated training schemes such as
Lagrangian Duality \cite{fioretto2020predicting,chatzos2020high} or
physics-informed neural network \cite{nellikkath2022physics}.  Graph
neural networks have been also considered in this context
\cite{donon2020deep,diehl2019warm,Owerko2020} for leveraging the power
system topology.  However, such direct approaches cannot scale to industry
size problems mainly because of the dimension
of the output space which is of very large scale.  To remedy this,
\emph{spatial decomposition} approaches \cite{chatzos2021spatial,mak2022learning} 
have been proposed to decompose the network in regions
and learn the mappings per region.

Besides the end-to-end learning approach, ML has
also helped optimization solve problems faster.
In \cite{deka2019learning}, the authors used an ML technique to identify an active set of
constraints in DC-OPF. Also in \cite{cengil2022learning}, ML is used to
identify a variable subset for accelerating an optimality-based bound
tightening algorithm \cite{sundar2018optimization} for the AC-OPF.

By definition, since learning the AC-OPF optimal solution is a regression task, 
the inference from the ML model will not be always correct. 
A variety of techniques have been used to remedy this limitation, 
including the use of warm-starts and power flows as post-processing.  
A Newton-based method is used to correct the active power generations and
voltage magnitudes at generator buses so that they satisfied the AC
power flow problem \cite{venzke2020inexact,chatzos2021spatial}. 
In \cite{taheri2022restoring}, the authors corrected voltages at
buses by minimizing the weighted least square of the inconsistency in
the AC power flow using the Newton-Raphson method.
In \cite{zamzam2020learning}, the active power injections
and voltage phasors are outputted directly from the ML model and
determine the other variables by solving a power flow.
Also, in \cite{baker2019learning} and \cite{dong2020smart}, the use of
the learning scheme was suggested to provide warm-start points for ACOPF solvers, 
but only presented the results on small test networks (up to the 300 bus system).
In \cite{pan2022deepopf}, the use of DNN-based learning method
for generating warm-start points was suggested. Their method was tested on a power
system with 2,000 buses, but no speed-up was reported with the learning-based 
warm-start point.

\tb{ In previous data-driven AC-OPF approaches, feature reduction has
  been considered for reducing the parameters to tune using techniques
  such as PCA \cite{lei2020data} or sensitivity analysis
  \cite{liu2021explicit}.  This work in contrast shows that the
  optimal solutions can be compressed to the low-rank representations
  defined by PCA without having any significant informational loss.  }
Traditional power network reduction techniques, such as Kron and Ward
reduction~\cite{ward49equivalent} techniques, have been widely used in
the power system industry for more than 70 years.  These techniques
focused on crafting simpler equivalent circuits to be used by system
operators, primarily for analysis.  More complex reduction
models~\cite{jang13line,caliskan12kron,nikolakakos18reduced,jiang20enhanced}
have also been developed recently.  While it is possible to use
classical reduction techniques to reduce the power systems before
learning, the resulting prediction model would only be able to predict
quantities on the reduced networks with potential accuracy issues.
The main focus of the paper is not on general network/grid
reduction techniques.  Instead, it focuses on devising a scalable
learning approach by reducing the number of trainable parameters.

In summary, most previous works for learning AC-OPF optimization proxies have not
considered industry-size power systems. {\em This paper shows that \CL{}
applies to large-scale power networks (up to 30,000 buses in the
experiments) and produces significant benefits in speeding-up AC-OPF
solvers through warm-starts.}

\section{Preliminaries}
\label{sec:acopf_ml_based}

This section formulates the AC-OPF problem and specifies the supervised learning studied in this paper. 

\subsection{AC-OPF Formulation}
\begin{model}[!t]
{\scriptsize
\caption{\tb{The AC-Optimal Power Flow (AC-OPF) Problem}}
\label{model:acopf_simple}
\begin{flalign}
&\Minimize_{\pg,\qg,\vm,\va} \sum_{i\in\cG}{c_i(\pgi)}\label{eq:acopf_obj_simple}\\
&\text{\textbf{subject to:}}\nonumber\\
&\theta_r= 0\label{eq:acopf_cnst_refva}\\ \setcounter{equation}{3}
&\pgimin\leq\pgi\leq\pgimax &\forall i\in\cG\label{eq:acopf_cnst_pgbound} \tag{\theequation a}\\
&\qgimin\leq\qgi\leq\qgimax &\forall i\in\cG\label{eq:acopf_cnst_qgbound} \tag{\theequation b}\\
&\underline{v}_i\leq v_i\leq\overline{v}_i &\forall i\in\cN\label{eq:acopf_cnst_vm}\\
\setcounter{equation}{5}
&S_{lij}\!=\!\left(Y_l^*\!-\!j\frac{b_l^c}{2}\right)\frac{v_i^2}{|T_l|^2}-Y_l^*\frac{V_iV_j^*}{T_l}              &\forall (lij)\!\in\!\cE\label{eq:acopf_cnst_flow1}\tag{\theequation a}\\
&S_{lji}\!=\!\left(Y_l^*\!-\!j\frac{b_l^c}{2}\right)v_i^2-Y_l^*\frac{V_i^*V_j}{T_l^*}              &\forall (lji)\!\in\!\cE^R\label{eq:acopf_cnst_flow2}\tag{\theequation b}\\
&\sum_{k\in\cG_i}S_k^g\!\!-\!\!S_i^d\!\!-\!\!Y_i^s|V_i|^2=\!\!\!\!\!\!\!\!\sum_{(lij)\in\cE\cup\cE^R}\!\!\!\!S_{lij} &\forall i\in\cN\label{eq:acopf_cnst_balance} \\
&|S_{lij}|\leq \bar{s}_l &\forall(lij)\!\in\!\cE\!\cup\!\cE^R\label{eq:acopf_cnst_thermal_limit}\\
&\underline{\Delta\theta}_l\leq \vai-\vaj\leq \overline{\Delta\theta}_l&\forall (lij)\!\in\!\cE\label{eq:acopf_cnst_angle_limit}
\end{flalign}
}
\end{model}

\tb{
Model~\ref{model:acopf_simple} presents the AC-OPF formulation 
\cite{babaeinejadsarookolaee2019power}.
The power network can be represented as a graph $(\cN,\cE)$ where 
$\cN$ denotes the set of bus indices containing generators and load units, 
and $\cE$ is the set of transmission line indices between two buses. 
$(lij)\in\cE$ where $l$ is a branch index connected from node $i$ to node $j$.
The set $\cE^R$ captures the reversed orientation of $\cE$, 
i.e., $(lji)\in\cE^R,\:\forall (lij)\in\cE$.
}

\tb{A generator output $S^g=\pg+j\qg$ is a complex number, where the
  real part $\pg$ is an active power generation (injection) and the
  imaginary part $\qg$ is a reactive power generation. An AC voltage
  $V=\vm\angle\va$ is represented by a voltage magnitude $\vm$ and a
  voltage angle $\va$.  The objective function
  ~\eqref{eq:acopf_obj_simple} minimizes the sum of quadratic cost
  functions $c_i(\cdot)$ with respect to the active power generations
  $\pgi,i\in\cG$.  At the reference bus $r$, the voltage angle is set
  to zero as defined in constraint~\eqref{eq:acopf_cnst_refva}.
  Constraints~\eqref{eq:acopf_cnst_pgbound},
  \eqref{eq:acopf_cnst_qgbound}, and \eqref{eq:acopf_cnst_vm} capture
  the bounds on variables $\pg$, $\qg$, and $\vm$, respectively.
  Constraints~\eqref{eq:acopf_cnst_flow1}, \eqref{eq:acopf_cnst_flow2}
  represent the complex power flow for each transmission line, which
  is governed by \emph{Ohm's law}. Here, $Y_l$, $b_l^c$, and $T_l$ are
  the series admittance, line charging susceptance, and transformer
  parameter at each branch $l$, respectively.
  Constraints~\eqref{eq:acopf_cnst_balance} ensures that, at each bus,
  the active and reactive power balance are satisfied.  Here, $Y^s$ is
  the bus shunt admittance and $\cG_i$ represents the set of generator
  indices attached to the bus $i$.
  Constraints~\eqref{eq:acopf_cnst_thermal_limit} capture the thermal
  limits and ensure that the apparent power does not exceed its limit
  $\bar{s}_{ij}$ for every transmission line.  }

\tb{
For the sake of simplicity, in what follows, the input configuration parameters 
and the optimal solution to the AC-OPF are denoted by $x$ and $y^*$, respectively.
}

\subsection{Supervised Learning}
The end-to-end learning approach in this paper consists in using
supervised learning to find a mapping from an input $x$ to an optimal
solution $y^*$ to AC-OPF. \tb{This work assumes that the set of
  commitment decisions and generator bids are predetermined.  This
  setting is common in prior end-to-end learning studies for
  AC-OPF (e.g.,
  \cite{zamzam2020learning,chatzos2020high,chatzos2021spatial}). Existing
  supervised learning schemes exploit the instance data
  $\{x_i,y^*_i\}$.  However, it often suffers from the high
  dimensionality of the output when dealing with power networks of
  industrial sizes.  Table~\ref{tab:case_spec} reports the
  specifications of the nine power networks from PGLib
  \cite{babaeinejadsarookolaee2019power} and a realistic
  version of the French system (denoted by \texttt{France\_2018}) used
  in the experiments.  The French system\footnote{\tb{Refer to
      \cite{chatzos2022data} for more details on this test case.}}
  uses the realistic grid topology of the French transmission system
  captured in 2018 with annual time series data fot the renewable
  generation capacity and load demand.}

\tb{ For the PGLib test cases, each instance has different active and
  reactive load demands $\pd$ and $\qd$, i.e., $x:=\{\pd,\qd\}$.  For
  \texttt{France\_2018}, the generation capacity of the renewable
  generators is also varying in addition to the load demands.  As
  such, for the PGLib cases, the dimensions of $x$ is
  $\dim\left(x\right)=2|\cL|$, and for \texttt{France\_2018}, the
  input of the mapping is increased to
  $x:=\{\pd,\qd,\{\overline{p}_i^{g} \}_{i \in \cG^r}\}$, where
  $\overline{p}$ represents the upper bounds of the active generation,
  and $\cG^r$ is the set of renewable generator indices.  Note that
  among 1890 generators in this system, there are 1609 renewable
  non-dispatchable generators including hydro, solar, and wind
  generators.  }

{\em From Table~\ref{tab:case_spec}, observe that the dimension of the
  output $y$ is much higher than that of $x$: this implies that the
  DNN for the OPF will necessitate an excessive number of trainable
  parameters.} It is the goal of this paper to propose a scalable
approach that mitigates this curse of dimensionality.

\tb{Note that one can recover the whole optimal solution estimates
  from the active generations and voltage magnitudes by solving a
  power flow problem as in \cite{zamzam2020learning}. This could be
  combined with the methods proposed herein to provide further
  reduction in the output space at the cost of a more costly training
  or inference procedure. However considering the full output space,
  i.e., $y:=\{\pg,\qg,\vm,\va\}$, makes it possible to use Lagrangian
  Duality \cite{fioretto2020predicting,chatzos2020high} or Primal-Dual
  Learning \cite{park2022self} frameworks to improve the learning
  procedure further.}

\section{Low Rank Representation of the AC-OPF Solution}
\label{sec:prelim}

\begin{table}[!t]
\centering
\small
\setlength{\tabcolsep}{2pt}
\begin{tabular}{@{}l|cccccccc@{}}
\toprule
Test case  & $|\cN|$ & $|\cG|$ & $|\cL|$ & $|\cE|$  &$\dim\!\left(x\right)$ &$\dim\!\left(y\right)$\\
\midrule
\texttt{300\_ieee}     & 300    & 69   & 201  & 411     & 402   & 738     \\
\texttt{793\_goc}      & 793    & 97   & 507  & 913     & 1014  & 1780    \\
\texttt{1354\_pegase}  & 1354   & 260  & 673  & 1991    & 1346  & 3228    \\
\texttt{3022\_goc}     & 3022   & 327  & 1574 & 4135    & 3148  & 6698    \\
\texttt{4917\_goc}     & 4917   & 567  & 2619 & 6726    & 5238  & 10968   \\
\texttt{6515\_rte}     & 6515   & 684  & 3673 & 9037    & 7346  & 14398   \\
\texttt{9241\_pegase}  & 9241   & 1445 & 4895 & 16049   & 9790  & 21372   \\
\texttt{13659\_pegase} & 13659  & 4092 & 5544 & 20467   & 11088 & 35502   \\
\texttt{30000\_goc}    & 30000  & 3526 & 10648& 35393   & 21296 & 67052   \\
\texttt{\tb{France\_2018}}  & 6708   & 1890 & 6262 & 8965    & 14133 & 17196   \\
\bottomrule
\end{tabular}
\caption{\tb{Specifications of the AC-OPF Test Cases.}}
\label{tab:case_spec}
\end{table}

This section presents data analysis to motivate \CL{}. Again,
Table~\ref{tab:case_spec} describes the test cases.  They range from
300 to 30,000 buses and from 69 to 3526 generators.  The table also
specifies the input and output dimensions of the learning problem: the
output dimension is large in sharp contrast to many classification
problems in computer vision for instance. The experiments in this
section are based on 20,000 instances for each test case: their
optimal solutions were obtained via {\sc Ipopt}
\cite{wachter2006implementation}.  \tb{To generate the instances for
  each test case in PGLib, the active loads were sampled from a
  truncated multivariate Gaussian distribution as}
\begin{equation}\label{eq:perturb_load}
    \pd \sim \mathcal{T}\mathcal{N}\left( p^d_0,\Sigma, (1-\mu)p^d_0, \mu p^d_0 \right),
\end{equation}
\tb{ where $p^d_0$ is the baseline active loads and $\Sigma$ is the
covariance matrix.  The element of the covariance matrix
$\Sigma_{ij}$ is defined using correlation coefficient $\rho$ as
$\Sigma_{ij} = \rho \sigma_i \sigma_j$ where $\sigma_i$ and
$\sigma_j$ are standard deviation of $\pd_i$ and $\pd_j$,
respectively.  Also, $\rho=1$ when $i=j$ and $\rho=0.5$ otherwise.
$\mu$ is set to $0.5$, meaning that the active load demands were set
to be perturbed by $\pm50\%$.  The reactive loads $\qd$ were sampled
from a uniform distribution ranging from $0.8$ to $1.0$ of the
baseline values.  }  This perturbation method follows the protocols
used in \cite{zamzam2020learning,donti2021dc3,park2022self}.  \tb{For
\texttt{France\_2018}, the experiments use the historical 2020 load
demand and renewable generations data at the 30-minute granularity,
which is publicly available at \cite{eco2mix}.  These historical
time series data are disaggregated spatially and interpolated to
have 5-minute granularity following the protocol introduced in
\cite{chatzos2022data}.} A PCA was performed on the 20,000 optimal
solutions for each test case, which led to a number of interesting
findings.

\begin{figure}[!t]
\centering
\includegraphics[width=.85\columnwidth]{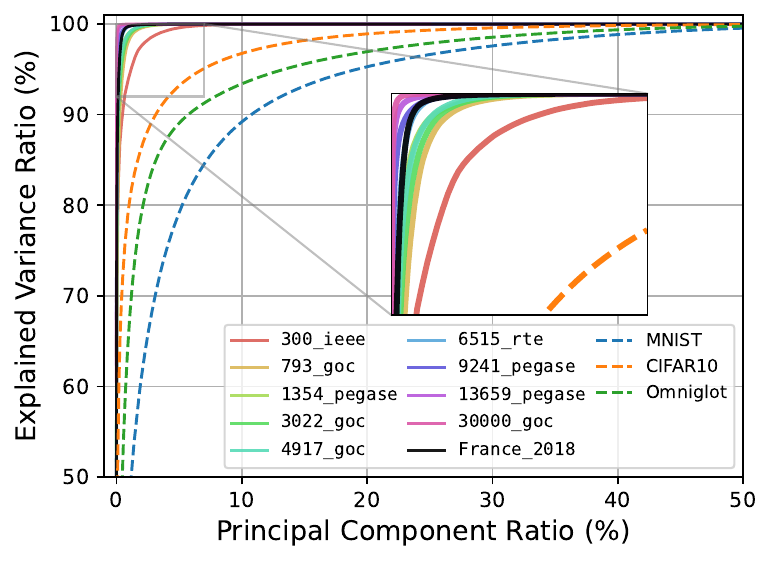}
\caption{\tb{Explained Variance Ratios of PCA For the Various Principal Component
Ratios for AC-OPF Instances (solid lines) and Natural Image Data (dashed lines).}}
\label{fig:pca}
\end{figure}

\begin{figure}[t!]
\centering
\includegraphics[width=.95\columnwidth]{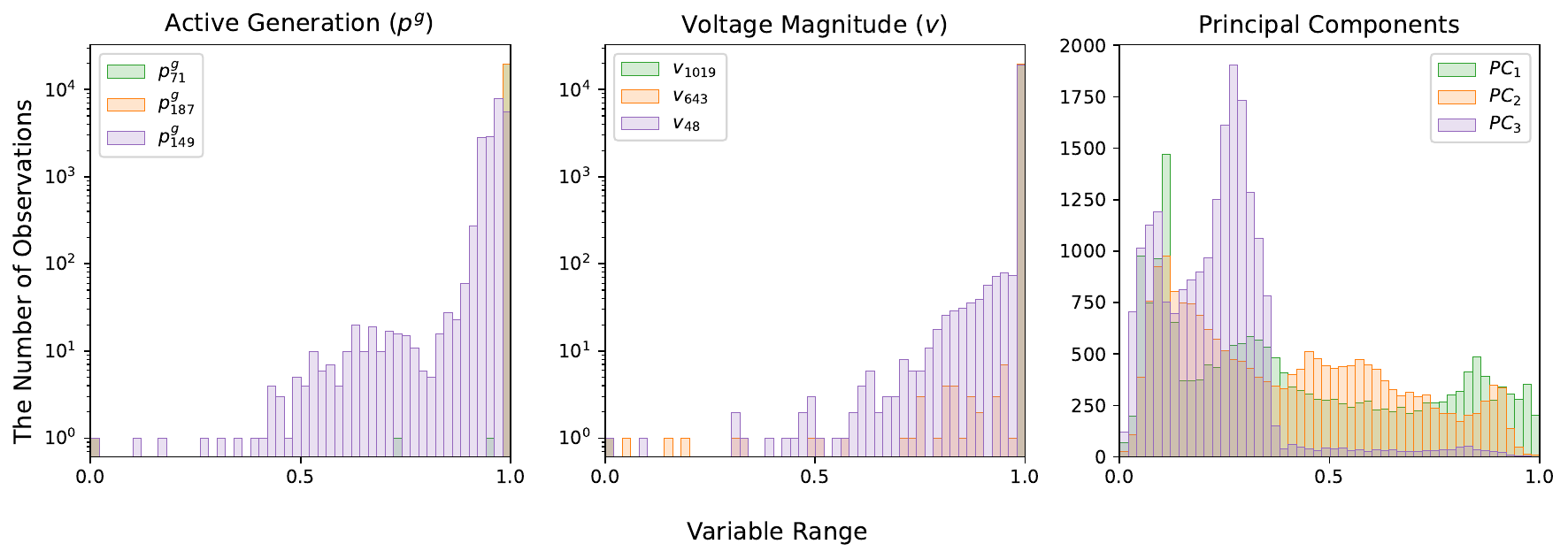}
\caption{Histograms of Active Generations (left), Voltage Magnitudes (middle), and Principal Components (right) of the 20,000 Optimal Solutions in \texttt{1354\_pegase}. Three largest components on average are illustrated. The x-axis is normalized to $[0,1]$ using the minimum and maximum values.}
\label{fig:hist}
\end{figure}

\paragraph{(Almost) Lossless Compression} A key observation of the analysis
is that PCA achieves an almost lossless compression with a few
principal components.  This is highlighted in Figure~\ref{fig:pca}
where the x-axis represents the principal component ratio (i.e., the
ratio of the number of the principal components in use to the
dimension of the optimal solution) and the y-axis represents the
explained variance ratio (i.e., the ratio of the cumulative sum of
eigenvalues of the principal components in descending order to the sum
of the all eigenvalues). The explained variance ratio is a proxy for
how much the information is preserved within the chosen low-rank
representation. For instance, the figure shows that $1\%$ of principal
components preserves the $99.92\%$ of information of the AC-OPF
optimal solutions for \texttt{13659\_pegase}. The detailed values are
shown in Table~\ref{tab:pca_ratio}, which highlights that the
compression is almost lossless with a 10\% principal component ratio.
This contrasts with data instances in computer vision, as exemplified
by the MNIST \cite{mnist}, CIFAR10 \cite{cifar10}, and Omniglot
\cite{omniglot} datasets. This result is encouraging: it suggests that
optimal solutions could be recovered with negligible losses when
learning takes place in a low-rank space of a few principal
components, potentially reducing the size of the mapping function
substantially.

\paragraph{Larger Test Cases Need Fewer Principal Components}
\tb{Figure~\ref{fig:pca} and Table~\ref{tab:pca_ratio} also highlight a
desirable trend: larger test cases need a lower ratio of principal components to
obtain the same level of explained variance ratio.}  
For instance, the explained
variance ratios of \texttt{13659\_pegase} and \texttt{30000\_goc} with
 a principal component ratio of $1\%$ are $99.92\%$ and $99.99\%$
respectively. In contrast, \texttt{300\_ieee}, the smallest test case, has $93.92\%$ 
of explained variance ratio. This observation shows that reducing 
the dimensionality through PCA is more effective for the bigger test cases
and will be used in deciding the learning architecture for different test cases.

\begin{table}[!t]
\centering
\small
\begin{tabular}{@{}lcccc@{}}
\toprule
           &\multicolumn{4}{c}{Principal Component Ratio} \\
            \cmidrule(lr){2-5}
Test case  & 1\% & 5\% & 10\% & 20\% \\
\midrule
\texttt{300\_ieee}          & 93.92 & 99.56 & 99.97  & 99.99\\
\texttt{793\_goc}           & 98.05 & 99.98 & 100.00 & 100.00\\
\texttt{1354\_pegase}       & 99.04 & 99.98 & 100.00 & 100.00\\ 
\texttt{3022\_goc}          & 98.64 & 99.99 & 100.00 & 100.00\\ 
\texttt{4917\_goc}          & 99.03 & 99.99 & 100.00 & 100.00\\ 
\texttt{6515\_rte}          & 99.74 & 99.99 & 100.00 & 100.00\\   
\texttt{9241\_pegase}       & 99.76 & 100.00 & 100.00 & 100.00\\ 
\texttt{13659\_pegase}      & 99.92 & 100.00 & 100.00 & 100.00\\ 
\texttt{30000\_goc}         & 99.99 & 100.00 & 100.00 & 100.00\\ 
\texttt{\tb{France\_2018}}  & 99.75 & 99.99 & 100.00 & 100.00\\\midrule   
MNIST                       & 44.45 & 79.12 & 89.17 & 95.27\\               
CIFAR10                     & 62.09 & 82.34 & 88.37 & 93.29\\               
Omniglot                    & 16.55 & 44.26 & 59.06 & 73.31\\               
\bottomrule
\end{tabular}
\caption{\tb{Explained Variance Ratios (\%) on Various Principal Component Ratios from 1\% to 20\%.}}
\label{tab:pca_ratio}
\end{table}

\paragraph{Smoother Distributions on the Principal Components}
Figure~\ref{fig:hist} provides some intuition for
why learning in the space of the principal components is
appealing. The figure shows the distributions of three active powers (left) and
voltage magnitudes (middle) in the optimal solutions for the
\texttt{1354\_pegase} test case.  The values are plotted in
log-scaled, highlighting the skewed nature of the active powers and
voltage magnitudes:
indeed, most values lie on their extreme limits. This has been observed
before, leading to the use of the
\emph{classification-then-regression} approach
\cite{chen2022learning}.  However, fortunately, Figure~\ref{fig:hist}
(right) shows that the distribution on the principal components is
well-posed: it is more convenient to learn the regression to the
low-rank space with a few principal components rather than to the 
original optimal solution space directly. As a result, the regression
learning in the space of
principal components should be easier than in the original space. 

\section{Compact Optimization Learning}
\label{sec:method}

\begin{figure}[t!]
\centering
\includegraphics[width=.90\columnwidth]{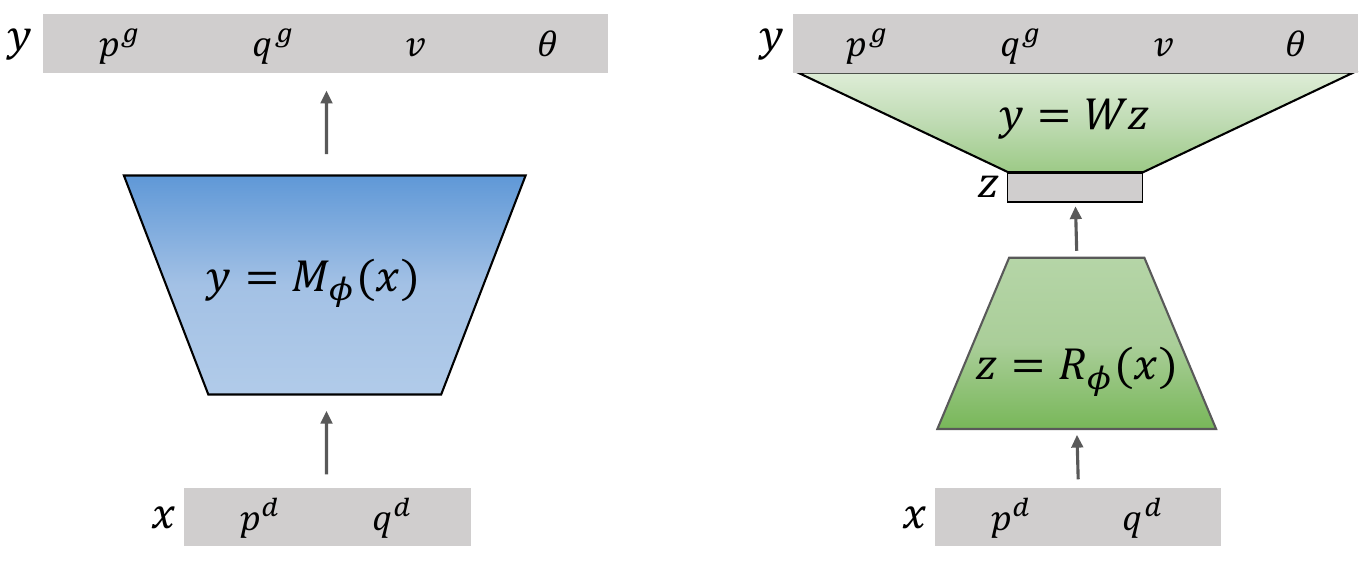}
\caption{Schematic View of the Plain Approach (left) and \CL{} (right).}
\label{fig:schematic}
\end{figure}

The findings in Section~\ref{sec:prelim} suggest a \CL{}
model whose outputs are in the subspace defined by the principal
components. This enables to decrease the number of trainable
parameters in the DNN-based mapping function, 
reducing the memory footprint significantly.

\subsection{Compact Learning}
The idea underlying \CL{} is to {\em jointly learn the principal
components and the mapping between the original inputs and the
outputs in the subspace of the principal components.}
Figure~\ref{fig:schematic} contrasts the overall architecture of the
\CL{} with the conventional plain learning approach. The plain
approach learns a mapping $y=M_\phi(x)$ from an input configuration
$x$ to an optimal solution $y$, where $\phi$ are the associated
trainable parameters. Since $y$ is high-dimensional, it is desirable
to have a large number of parameters to obtain accurate
solution estimates. \CL{} in contrast learns a mapping $z=R_\phi(x)$ from an input
configuration $x$ into an output $z$ in the subspace of principal
components before recovering an optimal solution prediction as $y = W
z$. The output dimension of $R_{\phi}$, $\dim\left(z\right)$, in \CL{}
should be substantially smaller than the dimension of $y$, making it
possible to reduce the number of trainable parameters.
In the upcoming sections, the way of learning $R$ and $W$ simultaneously 
is detailed.

\subsection{Learning the Principal Components}
\label{ssec:principal_components}

Let $p$ and $d$ be the dimensions of $z$ and $y$, respectively, 
i.e., $p=\dim\left(z\right)$ and $d=\dim\left(y\right)$. 
The goal of \CL{} is
to have $p$ substantially smaller than $d$, i.e., $p\ll d$. Matrix
$W\in\RR^{d\times p}$ is unitary, i.e., $W^\top W=I$, and its columns
are composed of the orthonormal principal components of the output
space associated with $p$ largest eigenvalues. It is obviously
possible to obtain $W$ through PCA, but this computation takes a
substantial amount memory when there are many instances.

Instead, \CL{} uses the Generalized Hebbian Algorithm (GHA), also
called Sanger's rule \cite{sanger1989optimal}, to learn $W$ in
a stochastic manner. GHA is specified in Algorithm~\ref{alg:GHA}. Using
the mini-batch of optimal solutions $\{y^{*(i)}\}$, it first updates the element-wise
running mean $\mu$ and variance $\sigma^2$ of optimal solution using the momentum
parameter $\beta$.  The momentum parameter $\beta$ is set to zero 
for the first iteration, and from the second iteration it is set to a 
value ranged $(0.0,1.0)$.
The optimal solution $y^{*(i)}$ is normalized
using the running mean and variance as shown in
line~\ref{algline:GHA:normalize} in Algorithm~\ref{alg:GHA}. GHA 
uses Gram-Schmidt process to find out the $\Delta W$ of the $p$
leading principal components, where $\mathcal{LT}[\cdot]$ represents the
lower triangular matrix.  Once $\Delta W$ is determined, $W$ is
updated by adding $\gamma\Delta W$, where $\gamma$ is the learning
rate that decreases as the epoch number $e$ increases, i.e.,
\begin{equation*}
    \gamma = \max\{\gamma_{\text{min}}, \gamma_{\text{init}}/(0.01e)\},
\end{equation*}
In the update rule, $\gamma_{\text{init}}$ and $\gamma_{\text{min}}$
are hyperparameters representing the initial and minimum learning
rate, respectively.

\tb{ One can simply use deterministic PCA to define $W$ instead of
  using GHA.  However this requires significant computing time before
  starting the training, especially when dealing with a large number
  of data instances of high dimensionality.  In contrast, GHA
  amortizes the computational effort over the training procedure
  leading to a better handling of the largest test cases.  }

\subsection{Training the Compact Learning Model}
\label{ssec:training}

\begin{algorithm}[!t]
\caption{Generalized Hebbian Algorithm,\\ \texttt{GHA}$(\{y^{(i)}\}_{i \in [B]},\!W,\!\mu,\!\sigma)$ }
\label{alg:GHA}
\textbf{Parameters}:\\
\-\hspace{0.1in}$\beta$: a momentum parameter\\
\-\hspace{0.1in}$\gamma$: a learning rate for updating $W$ \\
\-\hspace{0.1in}$\epsilon$: a small positive for numerical stability
\begin{algorithmic}[1] 
\State $m \gets \frac{1}{B}\sum_{i=1}^B y^{(i)}$
\State $s^2 \gets \frac{1}{B}(y^{(i)}-m)^2$
\State $\mu \gets \beta\mu + (1-\beta) m$ \Comment{update running mean}
\State $\sigma^2 \gets \beta\sigma^2 + (1-\beta) s^2$ \Comment{update running variance}
\State $\hat{y}^{(i)} \gets (y^{(i)}-\mu)/\sqrt{\sigma^2+\epsilon}$,  $\;\forall i \in [B]$ \Comment{normalize $y$}\label{algline:GHA:normalize}
\State $\Delta W \!=\! \frac{1}{B}\!\sum_{i=1}^B \!\left(\hat{y}^{(i)}\hat{y}^{(i)\top} W \!-\! W \mathcal{LT}[W^\top \hat{y}^{(i)} \hat{y}^{(i)\top} W]\right)$  
\State $W \gets W + \gamma\Delta W$ \Comment{update $W$}
\State \textbf{return} $W$, $\mu$, $\sigma$.
\end{algorithmic}
\end{algorithm}

\begin{algorithm}[!t]
\caption{\CL{}}
\label{alg:compact_learning}
\begin{algorithmic}[1] 
\For{$k$=1\dots}
\State Sample $\{(x^{(i)},y^{*(i)}\}_{i \in [B]}$ from $\dataset$
\State $W,\mu,\sigma \gets$ \texttt{GHA}$(\{y^{*(i)}\}_{i \in [B]},W,\mu,\sigma)$ \label{algline:compact:GHA}
\State $z^{(i)} = R_{\phi}(x^{(i)})$, $\;\forall i \in [B]$
\State $y^{(i)} = \sqrt{\sigma^2\!+\!\epsilon} \,W z^{(i)}\!+\!\mu$, $\;\forall i \in [B]$ \label{algline:compact:denorm}
\State Update $\phi$ with $\cL=\frac{1}{B}\sum_{i=1}^B\norm{{y^{(i)}}-{y^{*(i)}}}_1$
\EndFor{}
\end{algorithmic}
\end{algorithm}

The training process of \CL{} is summarized in
Algorithm~\ref{alg:compact_learning} {\em which jointly learns the principal components $W$ and
the mapping $R$.} Each iteration samples a mini-batch of size $B$
and applies the GHA Algorithm~\ref{alg:GHA} (line
\ref{algline:compact:GHA} in Algorithm~\ref{alg:compact_learning}).  
The mapping $R$ then produces $z^{(i)}$ for
each input $x^{(i)}$ and the output $y^{(i)}$ is obtained through the mapping
$W z^{(i)}$ and a denormalization step (line
\ref{algline:compact:denorm} in Algorithm~\ref{alg:compact_learning}). 
Finally, the parameters $\phi$
associated with $R$ are updated using backpropagation from the loss
function $\cL$, which captures the mean absolute error between the
prediction $y^{(i)}$ and the optimal solution $y^{*(i)}$ (ground truth).

\subsection{Restoring Feasibility}
\label{alg:postprocessing}

The predictions from \CL{} may violate the physical and engineering
constraints of the AC-OPF. Some applications are required to address these
infeasibilities and this study considers two post-processing
methods for this purpose: (1) solving the power flow problem; and (2)
warm-starting the exact AC-OPF solver.

\subsubsection{Power Flow}

The power flow problem, seeded with a prediction from the \CL{} model,
restores the feasibility of the physical constraints.  Formally, the
power flow problem can be formulated as:
\begin{equation}
\label{eq:pf_formulation}
\begin{aligned}
&&&\text{\textbf{find}} && \pg, \qg, \vm, \va,\\
&&&\text{\textbf{subject to }} && \text{Eq.~\eqref{eq:acopf_cnst_flow1}, \eqref{eq:acopf_cnst_flow2}},\\ 
&&&                            && 
\text{Eq.~\eqref{eq:acopf_cnst_balance}}.
\end{aligned}
\end{equation}
In the power flow problem, like in \cite{chatzos2021spatial}, the active power
injections and voltage magnitudes at the PV buses are fixed to the
predictions. The power flow problem
can then be solved by the Newton method to satisfy the physical
constraints, i.e., {\em Ohm's law}
(Eq.~\eqref{eq:acopf_cnst_flow1} and
\eqref{eq:acopf_cnst_flow2}) and \emph{Kirchhoff’s Current Law}
(Eq.~\eqref{eq:acopf_cnst_balance}). Finding a solution to the PF
problem typically takes significantly less time than solving the
AC-OPF. However, the solution may violate some of the engineering
constraints of the AC-OPF.

\subsubsection{Warm-Starting the AC-OPF solver}

It is possible to remove all infeasibilities by warm-starting an
AC-OPF solver with the \CL{} predictions. This study uses the
primal-dual interior point algorithm {\sc Ipopt} as a solver, which is
a standard tool for solving AC-OPF problems
\cite{babaeinejadsarookolaee2019power,gopinath2022benchmarking}.
Moreover, warm-starts for the primal-dual interior point algorithm
seem to benefit significantly from dual initial points. Hence, the
\CL{} model was generalized to predict dual optimal values for all
constraints in addition to the primal optimal solutions. {\sc Ipopt}
is then warm-started with both primal and dual predictions to obtain
an optimal solution.

\section{Computational Experiments}
\label{sec:exp}

\subsection{The Experiment Setting}
\label{ssec:exp_setting}

\tb{The performance of \CL{} is demonstrated using nine test cases
  from PGLIB v21.07 and the realistic version of the French power
  system (denoted as \texttt{France\_2018}) as described in
  Table~\ref{tab:case_spec}.  A total of 52,000 instances were
  generated.  50,000 instances are used for training and the remaining
  2,000 instances are tested for reporting the performance results.
  For the PGLib cases, these instances were obtained by perturbing the
  load demands as Eq~\eqref{eq:perturb_load} in
  Section~\ref{sec:prelim}: } \tb{ For \texttt{France\_2018}, 100,000
  instances for training are generated by perturbing the instances in
  September. Specifically, the upper bounds of active generation of
  the wind and hydro renewable generators are perturbed by replacing
  with the samples from
  $\mathcal{N}(\overline{p}^g_i,0.2(\overline{p}^g_i-\underline{p}^g_i))$.
  Also, the upper bounds of solar generators and load demands are
  perturbed by multiplying factor sampled from multivariate Gaussian
  $\mathcal{N}(\mathbf{1}, \Sigma)$ where $\Sigma$ is based on the
  correlation coefficient of $0.8$.  Those perturbed upper bound of
  active generation is ensured to be greater than or equal to the
  lower bound of it.  Also, 2000 realistic test instances are
  extracted from the instances in October for reporting the
  performance. As such, for this test case, the distribution of test
  instances is not necessarily the same as that of the training
  instances.  Note that accurate renewable forecasting would improve
  the quality of the training instances, but this experiment setting
  for \texttt{France\_2018} should provide realistic and difficult
  circumstances where the model may be deployed for real operations.
} The instances were solved using Pyomo \cite{hart2017pyomo} and {\sc
  Ipopt} v3.12 \cite{wachter2006implementation} with the HSL ma27
linear solver.

The performance of \CL{} is compared with the plain approach that
directly outputs the optimal solution.  As in
\cite{chatzos2021spatial}, four fully-connected layers followed by
ReLU activations are used for the mapping functions for both \CL{} and
plain learning approaches.  For the plain approach, two distinct
models are experimented; the first model, which is named {\sc
  Plain-Large}, has $d$ hidden nodes for each fully-connected layer
(where $d$ is the output dimension). The second baseline, which is
named {\sc Plain-Small}, has the $p$ hidden nodes for each layer
(where $p$ is the number of principal components considered in \CL{}).
The \CL{} model has the same number of weight parameters as {\sc
  Plain-Small}, but the last layer of the \CL{} model is learned
through GHA.  Indeed, {\sc Plain-Small} has an encoder-decoder
structure as the number of the hidden nodes is smaller than that of
the output. \tb{ The ratio of $p$ to $d$, which is also called
  principal component ratio, is set to $5\%$ for six smaller test
  cases (up to \texttt{6515\_rte}) and to $1\%$ for four bigger cases.
  Mini-batch of $64$ instances is used, and the maximum epoch is set
  to $1,000$.  }  The models are trained using the Adam optimizer
\cite{kingma2014adam} with a learning rate of
$1\mathrm{e}{\textrm{-}4}$, which is decreased at $900$ epochs by
$0.1$.  The overall implementation used PyTorch and the models were
trained on a machine with a NVIDIA Tesla V100 GPU and Intel Xeon
2.7GHz. For GHA (Algorithm~\ref{alg:GHA}), the momentum parameter
$\beta$ is set to $0.9999$ from the second iteration.  The initial and
minimum learning rate ($\gamma_{\text{init}}$ and
$\gamma_{\text{min}}$) are set to $1\mathrm{e}{\textrm{-}4}$, and
$1\mathrm{e}{\textrm{-}8}$, respectively.  The parameter $\epsilon$ to
prevent ill-conditioning is set to $1\mathrm{e}{\textrm{-}8}$.

\subsection{Learning Performance}
\label{ssec:exp_result_performance}

Table~\ref{tab:performance} reports the accuracy of the models for
predicting optimal solutions. Five distinct
models with randomly initialized trainable parameters per method were
trained: the results report the average results and the standard
deviations (in parenthesis).
The table shows the averaged value of optimality gaps and maximum constraint violations on the 2,000 test instances.
The optimality gap is calculated as
$100\times|\frac{f(y)-f(y^*)}{f(y^*)}|$ where $f(y)$ is the objective function~\eqref{eq:acopf_obj_simple}. 
It also reports the maximum
constraint violations (in per unit), \tb{which is computed as}
\begin{equation}
    \max\{\max_{i}\{\max\{g_i(y),0.\}, \max_i\{|h_i(y)|\}\},
\end{equation}
\tb{ where $g_i(y)$ and $h_i(y)$ are, respectively, the inequality and
  equality constraints enumerated in Model~\ref{model:acopf_simple}.
} The first two sets of columns represent the performance of the plain
approaches.  Note that, because of the high dimensionality of the
output and the limited GPU memory, {\sc Plain-Large} is only
applicable to the four smaller test cases (up to \texttt{3022\_goc}):
this is the limitation that motivated this study.  When comparing the
two plain models, {\sc Plain-Large} performs better than {\sc
  Plain-Small} as {\sc Plain-Small} trades off the accuracy for
scalability.  \CL{} almost always performs better than the two plain
approaches. In particular, it produces predictions with significantly
fewer violations (sometimes by an order of magnitude), while also
delivering smaller optimality gaps on the larger test cases. 

Table~\ref{tab:params} shows the number of trainable parameters of the
\CL{} model and the plain approaches.  The architectures of
\CL{} and {\sc Plain-Small} are exactly the same, except for the last
layer: hence the number of trainable parameters of \CL{} is smaller
than those for {\sc Plain-Small} by the dimension of $W$. 
Table~\ref{tab:params} clearly shows that \CL{} has the
smallest number of trainable parameters. Overall,
Table~\ref{tab:performance} and Table~\ref{tab:params} indicate that
\CL{} provides an accurate and scalable approach to predict AC-OPF solutions.

\begin{table*}[!t]
\centering
\small
\begin{tabular}{@{}l|cccccc@{}}
\toprule
              & \multicolumn{2}{c}{\sc{Plain-Small}} & \multicolumn{2}{c}{\sc{Plain-Large}} & \multicolumn{2}{c}{\CL{} (proposed)} \\
              \cmidrule(lr){2-3}\cmidrule(lr){4-5}\cmidrule(lr){6-7}
Test case               & Opt. Gap(\%) & Viol. & Opt. Gap(\%) & Viol. & Opt. Gap(\%) & Viol. \\
\midrule
\texttt{300\_ieee}      & 0.9886(0.0287) & 3.7282(0.0758) & \bb{0.1804}(0.0087) & 0.9256(0.0117) & 0.1859(0.0079)      & \bb{0.5509}(0.0025) \\
\texttt{793\_goc}       & 0.0579(0.0067) & 3.1791(0.0985) & \bb{0.0182}(0.0013) & 0.5825(0.0046) & 0.0305(0.0046)      & \bb{0.2768}(0.0033) \\
\texttt{1354\_pegase}   & 0.1494(0.0142) & 6.6858(0.1248) & \bb{0.0711}(0.0094) & 3.0360(0.1221) & 0.0942(0.0089)      & \bb{0.5044}(0.0081) \\
\texttt{3022\_goc}      & 0.0745(0.0105) & 3.2394(0.0846) & 0.0588(0.0094)      & 1.5837(0.1432) & \bb{0.0485}(0.0076) & \bb{0.3906}(0.0185) \\
\texttt{4917\_goc}      & 0.0852(0.0118) & 2.6590(0.1006) & -                   & -              & \bb{0.0527}(0.0087) & \bb{0.5456}(0.0276) \\
\texttt{6515\_rte}      & 0.7232(0.0364) & 5.8648(0.2185) & -                   & -              & \bb{0.3065}(0.0177) & \bb{0.7519}(0.0290) \\
\texttt{9241\_pegase}   & 0.1521(0.0138) & 6.7379(0.1980) & -                   & -              & \bb{0.1344}(0.0097) & \bb{1.1791}(0.0725) \\
\texttt{13659\_pegase}  & 0.0910(0.0057) & 4.8141(0.0685) & -                   & -              & \bb{0.0745}(0.0046) & \bb{1.2915}(0.0497)\\
\texttt{30000\_goc}     & 0.2296(0.0224) & 4.7172(0.0678) & -                   & -              & \bb{0.1091}(0.0112) & \bb{0.0770}(0.0188)\\
\texttt{\tb{France\_2018}} & 1.4872(0.1352) & 47.4012(3.4524) & -                   & -              & \bb{1.2869}(0.0845) & \bb{2.1041}(0.2218)\\
\bottomrule       
\end{tabular}
\caption{\tb{Performance Results of \CL{} (Proposed) and Conventional Plain Learning 
  Approaches (Baselines). Std. dev. in parenthesis is evaluated across five
  independent runs. \emph{Viol.}: the mean value of the maximum
  constraint violations (in per unit) on the test instances. The best
  optimality gap (\emph{Opt. Gap}) and maximum violation values are in
  bold.}}
\label{tab:performance}
\end{table*}

\begin{table}[!b]
\centering
\small
\begin{tabular}{@{}l|ccc@{}}
\toprule
Test case      & \sc{Plain-Small} & \sc{Plain-Large} & \compact \\
\midrule
\texttt{300}   & 0.045            & 2.479            & 0.019 \\
\texttt{793}   & 0.274            & 14.487           & 0.122 \\
\texttt{1354}  & 0.818            & 46.041           & 0.321 \\
\texttt{3022}  & 3.631            & 200.572          & 1.499 \\
\texttt{4917}  & 9.795            & -                & 4.074 \\
\texttt{6515}  & 17.202           & -                & 7.353 \\
\texttt{9241}  & 6.796            & -                & 2.268 \\
\texttt{13659} & 16.954           & -                & 4.442 \\
\texttt{30000} & 60.610           & -                & 16.067 \\
\tb{\texttt{France}}& 51.612           & -           & 30.510 \\
\bottomrule       
\end{tabular}
\caption{The Number of Trainable Parameters (in Millions) in the Models Trained by \CL{} and Plain Approaches.}
\label{tab:params}
\end{table}

\subsection{Post-processing: Power Flow}
\label{ssec:exp_result_pf}
One model among five trained models was randomly chosen for testing
post-processing approaches and the results were evaluated on the 
same 2,000 test instances.
Table~\ref{tab:pf_result} reports the constraint violations after
applying the power flow model seeded with the predictions from 
the \CL{} and {\sc Plain-Small} models.
The table also reports the time to solve the power flow problem.  
The results show that
\CL{} produces power flow solutions with the smallest constraint violations,
sometimes by an order of magnitude. 
The results of \CL{} are particularly impressive because the majority of the constraints are satisfied after applying the power flow.
Note also that the power flow is
fast enough to be used during real-time operations, opening
interesting avenues for the use of learning and optimization in
practice.

\subsection{Post-processing: Warm-start}
\label{ssec:exp_result_warmstart}

\begin{figure*}[!h]
\centering
\includegraphics[width=.7\textwidth]{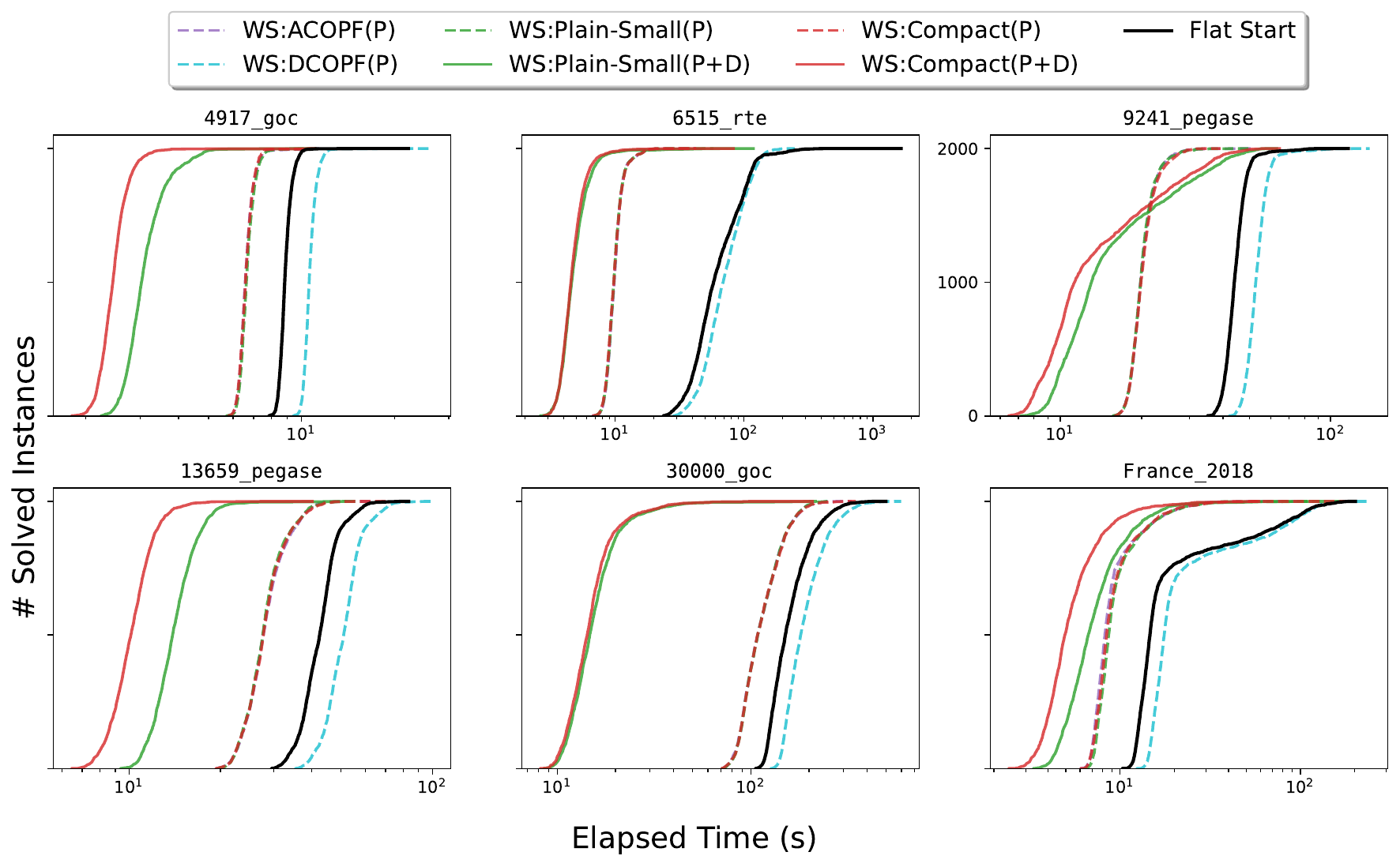}
\caption{\tb{The Number of AC-OPF Instances Solved Within Various Elapsed Time Limits by Warm-Start (WS) and Flat Start.}}
\label{fig:warmstart}
\end{figure*}

Table \ref{tab:warmstart} and Figure \ref{fig:warmstart} report the results for warm-starts. 
The proposed warm-starting approach, WS:\compact(P+D), is compared with the following warm-starting strategies:
\begin{itemize}
    \item Flat Start: $\pg$, $\qg$, $\vm$ are started with their minimum values, and initial $\va$ is set to zero. This is a default setting without warm-start.
    \item WS:DC-OPF(P): Motivated from \cite{venzke2020inexact}, the primal solution of the DC-OPF is used as a warm-starting point for solving AC-OPF.
    \item WS:AC-OPF(P): The primal solution of the AC-OPF is used as a warm-starting point for solving AC-OPF again.
    \item {WS:\sc Plain-Small(Large)(P)}: The primal predictions from the plain approaches are used as warm-starting points.
    \item {WS:\sc Plain-Small(Large)(P+D)}: The primal and dual predictions from the plain approaches are used as warm-starting points.
    \item WS:\compact(P): The primal predictions from \CL{} are used as warm-starting points.
    \item WS:\compact(P+D): The proposed warm-starting approach. The primal and dual predictions from \CL{} are used as warm-starting points.      
\end{itemize}

\begin{table*}[!t]
\centering
\small
\begin{tabular}{@{}l|ccccc|ccccc@{}}
\toprule
              & \multicolumn{5}{c}{\sc{Plain-Small}} & \multicolumn{5}{c}{\CL{}} \\
              \cmidrule(lr){2-6}\cmidrule(lr){7-11}
              & \multicolumn{2}{c}{Eq.(\ref{eq:acopf_cnst_pgbound}-\ref{eq:acopf_cnst_vm})} & \multicolumn{2}{c}{Eq.~\eqref{eq:acopf_cnst_thermal_limit}} 
              & Time
              & \multicolumn{2}{c}{Eq.(\ref{eq:acopf_cnst_pgbound}-\ref{eq:acopf_cnst_vm})} & \multicolumn{2}{c}{Eq.~\eqref{eq:acopf_cnst_thermal_limit}} 
              & Time \\
Test case     & Viol.(p.u.) & Sat(\%)
              & Viol.(MVA) & Sat(\%)
              & sec.
              & Viol.(p.u.) & Sat(\%)
              & Viol.(MVA) & Sat(\%)
              & sec. 
              \\
\midrule
\texttt{4917\_goc}   & 0.1148 & 99.77 & 13.5821 & 99.47 & 1.07 & {\bf 0.0739} & 99.88 & {\bf 2.3134}& 99.50 & 1.15 \\
\texttt{6515\_rte}   & 0.2220 & 99.80 & 6.0533  & 99.92 & 1.54 & {\bf 0.0841} & 99.92 & {\bf 1.5261}& 99.95 & 1.58 \\
\texttt{9241\_pegase}   & 0.1338 &99.47 & 10.3893 & 99.98 & 3.63 & {\bf 0.0893} &99.82& {\bf 2.4869}& 99.97 & 3.58 \\
\texttt{13659\_pegase}  & 0.2192 & 99.85 & 14.5279 & 99.99 & 7.11 & {\bf 0.0852} & 99.97 & {\bf 3.1203}& 100.00 & 7.01 \\
\texttt{30000\_goc}  & 0.0249 & 99.97 & 4.5279 & 100.00 & 10.27 & {\bf 0.0192}& 100.00 & {\bf 1.1958} & 100.00 & 10.11 \\
\texttt{France\_2018} & 0.4073 & 99.98 & 11.5033 & 99.02 & 1.86 & \bf{0.3788} & 99.98 & \bf{3.2053} & 99.08 & 1.93 \\
\bottomrule
\end{tabular}
\caption{\tb{Averaged Maximum Violations and Ratio of Satisfied constraints (\%) after Applying the Power Flow to AC-OPF Problems with $>$4,000 Buses.}}
\label{tab:pf_result}
\end{table*}

\begin{table*}[!t]
\centering
\small
\setlength{\tabcolsep}{2.5pt}
\begin{tabular}{lcc|rrrrrrr}
\toprule
                        & Primal & Dual & \multicolumn{1}{c}{\texttt{4917}} & \multicolumn{1}{c}{\texttt{6515}} & \multicolumn{1}{c}{\texttt{9241}} & \multicolumn{1}{c}{\texttt{13659}} & \multicolumn{1}{c}{\texttt{30000}} & \multicolumn{1}{c}{\texttt{France}} \\ \midrule
Flat Start              &        &      & \multicolumn{1}{c}{8.90s}       & \multicolumn{1}{c}{69.90s}       & \multicolumn{1}{c}{44.67s}       & \multicolumn{1}{c}{42.79s}        & \multicolumn{1}{c}{163.48s}       & \multicolumn{1}{c}{26.78s}        \\ \midrule
WS:AC-OPF(P) & \multirow{5}{*}{\cmark{}} & \multirow{5}{*}{\xmark{}} & 6.64s(1.34\texttimes) & 10.10s(7.06\texttimes) & 20.27s(2.22\texttimes) & 28.45s(1.53\texttimes) & 116.33s(1.43\texttimes) & 9.66s(2.62\texttimes)\\
WS:DC-OPF(P)            & & & 10.68s(0.83\texttimes) & 75.55s(0.90\texttimes) & 53.59s(0.83\texttimes) & 51.35s(0.83\texttimes) & 196.17s(0.83\texttimes) & 30.11s(0.85\texttimes)\\
WS:\sc{Plain-Small(P)}  & & & 6.69s(1.33\texttimes) & 10.06s(7.09\texttimes) & 20.24s(2.22\texttimes) & 28.24s(1.54\texttimes) & 116.23s(1.43\texttimes) & 10.04s(2.54\texttimes)\\
WS:\compact(P)          & & & 6.63s(1.34\texttimes) & 10.05s(7.10\texttimes) & 20.38s(2.21\texttimes) & 28.41s(1.53\texttimes) & 116.65s(1.43\texttimes) & 9.83s(2.61\texttimes)\\
\midrule
WS:\sc{Plain-Small(P+D)}& \multirow{3}{*}{\cmark{}} & \multirow{3}{*}{\cmark{}} & 3.16s(2.89\texttimes)  & 5.12s(14.79\texttimes) & 17.31s(3.17\texttimes) & 14.31s(3.05\texttimes) & 16.54s(10.98\texttimes) & 7.98s(3.65\texttimes)\\
WS:\compact(P+D)        & & & 2.49s(\bf{3.62}\texttimes) & 4.86s(\bf{15.26}\texttimes) & 15.59s(\bf{3.58}\texttimes) & 10.32s(\bf{4.23}\texttimes) & 15.66s(\bf{11.43}\texttimes) & 5.90s(\bf{4.84}\texttimes)\\
\bottomrule
\end{tabular}
\caption{\tb{Averaged Elapsed Times (s) to Solve the AC-OPF Problems with $>$4,000 Buses and Averaged Speed-up Factor to Flat Start. The Best Values are in Bold.}}
\label{tab:warmstart}
\end{table*}

\noindent
Obviously, WS:DC-OPF(P) and WS:AC-OPF(P) need to solve the first problem 
to obtain the warm-starting points. For those, the time taken to solve the first problem 
is excluded in the reported elapsed time performance. 
Note that WS:AC-OPF(P) gives a virtual upper bound of the speed-up 
for primal-only warm-start for the primal-dual interior point algorithm.
Also note that except for DC-OPF, the experiment does
not include other convex relaxations of AC-OPF (e.g., the quadratic convex
relaxation \cite{coffrin2015qc} and the semidefinite programming
relaxations \cite{bai2008semidefinite}), since solving those relaxed
problems takes significant time. 

To predict dual solutions, an additional
mapping function consisting of four fully-connected layers is trained for \CL{}
and the plain approaches. The sizes of these networks are the
same as those for the primal solutions. When using both the primal and dual
warm-starting, the initial barrier parameter of {\sc Ipopt} is set to
$1\mathrm{e}{\textrm{-}6}$ because the initial warm-starting point is
closer to the optimal than the flat start.  The convergence tolerance is
set to $1\mathrm{e}{\textrm{-}4}$ for all cases.

Table \ref{tab:warmstart} reports the elapsed times and the
corresponding speed-up ratio for the warm-start approaches.  The first
key observation is that it is critical to use both primal and dual
warm-starts: only using the primal predictions is not effective in
reducing the computation times of {\sc Ipopt}.  This is not too
surprising given the implementation of interior-point methods.
Primal-dual warm-starts however produce significant benefits.
WS:\compact(P+D) produces the best results for all test cases.  In
particular, it yields a speed-up of $15.26$\texttimes{} for
\texttt{6515\_rte}.  \tb{ Also, even for the realistic French system
  (\texttt{France\_2018}), WS:\compact(P+D) gives a speed-up of
  $4.84$\texttimes{}.} This is significant given the realism of this
test case and highlights the potential of the combination of \CL{} and
optimization to deploy AC-OPF in real operations. Observe also that
WS:\compact(P+D) strongly dominates the other approaches, including
its (WS:{\sc Plain-Small(P+D)} counterpart.  This was anticipated by
its prediction errors reported earlier. Figure \ref{fig:warmstart}
depicts these results visually: it plots the number of AC-OPF
instances solved over time by the various warm-starting methods.  The
plot clearly demonstrates the benefits of \CL{} for predicting both
primal and dual solutions.

\section{Conclusions}
\label{sec:conclusions}

This paper has proposed \CL{}, a novel approach to predict optimal
solutions to industry size OPF problems. \CL{} was motivated by the
lack of scalability of existing ML methods for this
task.  This difficulty stems from the dimension of the output space,
which is large-scale in industry size AC-OPF problems.  To address this issue,
\CL{} applies the PCA on the output space
and learns in the subspace of a few leading principal components. It
then combines this learning step with the GHA to learn the principal 
components, which is then used to transform the predictions back into
the output space.  
Experimental results on industry size OPF problems show that
\CL{} is more accurate than existing approaches both in terms of
optimality gaps and constraint violations, sometimes by an order of
magnitude.

The paper also shows that the predictions can be used to accelerate
the AC-OPF. In particular, the results show that
the power flow problems seeded by the \CL{} predictions have
significantly fewer violations of the engineering constraints (while
satisfying the physical constraints) for systems with up to 30,000
buses. Moreover, and even more interestingly, \CL{} can be used to
warm-start an OPF solver with optimal predictions for both the primal
and dual variables. The results show that \CL{} can produce
significant speed-ups.

{\em Together these results indicate that \CL{} is the first method to
  produce high-quality predictions for the industry size OPFs that
  translate to significant practical benefits.}  There are also many
opportunities for future research.  \CL{} is general-purpose and can
be applied to other problems with large output space, which is
typically the case in optimization. Nonlinear compression through
autoencoder structure can be also considered for general-purpose
optimization learning.  \tb{ On OPF problems, the performance of \CL{}
  can be improved by adopting concepts from Lagrangian Duality,
  physics-informed networks, and different types of backbone
  architectures.  The dual solution learning is particularly
  challenging in the experiments given its high dimensionality, and it
  would be interesting to study how it could be improved and
  simplified. Also, \CL{} can be extended to learning solutions to
  more challenging problems such as Optimal Transmission Switching and
  Security-Constrained OPF in power system applications.  }

\section*{Acknowledgments}
\tb{The authors thank Minas Chatzos for the discussions regarding the
  implementation of the power flow problem.  We also would like to
  express our gratitude to the anonymous reviewers whose insightful
  comments have greatly improved the quality of this paper. This
  research is partly supported by NSF Awards 2007095 and 2112533.}

\bibliographystyle{IEEEtran} 
\bibliography{refs}

\end{document}